\documentclass[a4paper]{article}

\usepackage{INTERSPEECH_v2}

\title{Improving Speaker-Independent Lipreading with Domain-Adversarial Training}
\name{Michael Wand, J\"{u}rgen Schmidhuber}
\address{
The Swiss AI Lab IDSIA, USI \& SUPSI, Manno-Lugano, Switzerland
}
\email{michael@idsia.ch, juergen@idsia.ch}

\usepackage{array}

\usepackage{xspace}
\newcommand*{\eg}{e.g.\@\xspace}
\newcommand*{\ie}{i.e.\@\xspace}

\usepackage{url}
\usepackage{multirow,makecell}

\newcommand{\tpm}{$\pm$}

\newcommand{\GRIDPictureExamples}[1]{%
\begin{figure}[#1]
\begin{center}
\centerline{\includegraphics[width=0.4\columnwidth]{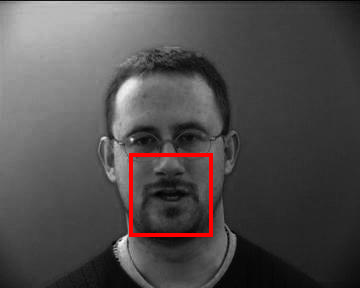}\includegraphics[width=0.4\columnwidth]{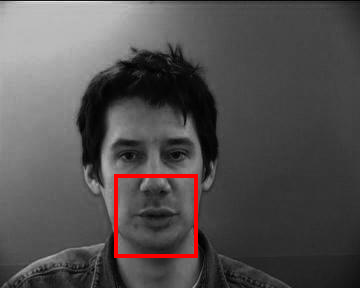}}
\caption{Two randomly chosen example frames from the GRID corpus with highlighted mouth area.}
\label{fig:GRIDExample}
\end{center}
\end{figure}}

\newcommand{\CurvesWithAdv}[1]{%
\begin{figure}[#1]
\begin{center}
\includegraphics[width=1.0\columnwidth]{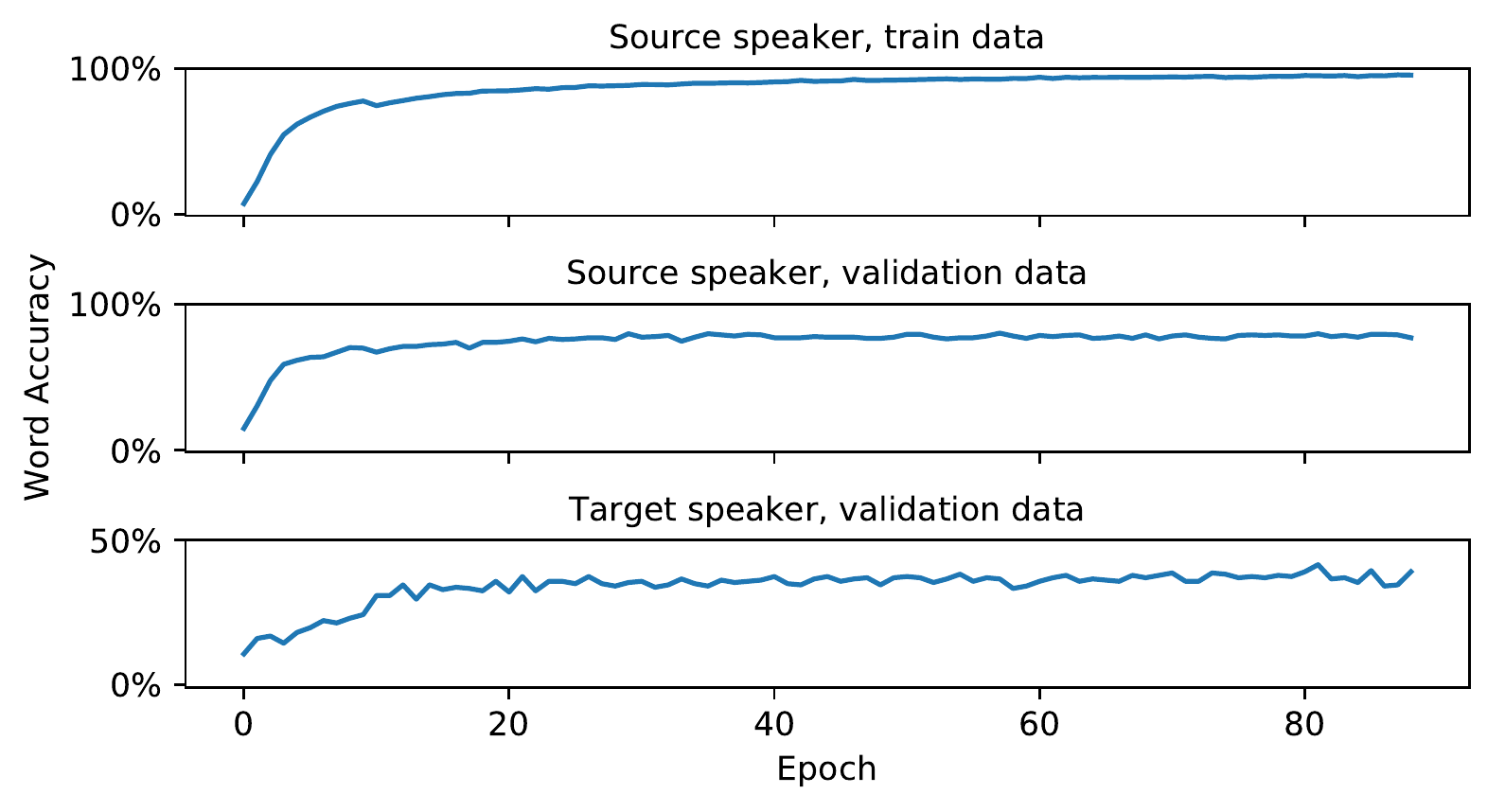}
\caption{Accuracy vs.~epoch on different data sets \textit{with} adversarial training, for speaker pair s5 $\rightarrow$ s6. Note that
the target accuracy shows a substantial rise at epoch 10, where adversarial training sets in.}
\label{fig:CurvesWithAdv}
\end{center}
\end{figure}}

\newcommand{\OverviewPairs}[1]{%
  \begin{figure}[#1]
    \includegraphics[width=7cm]{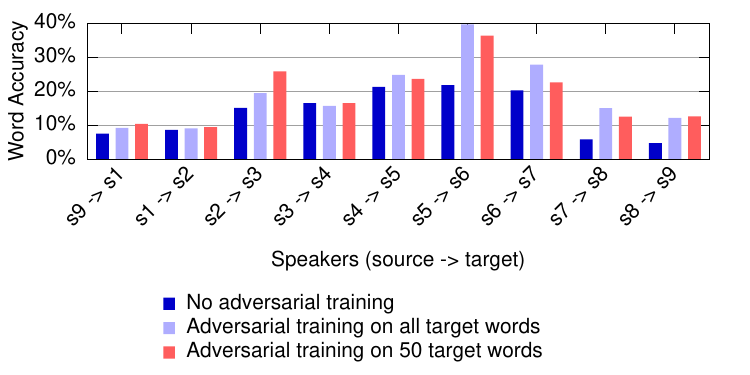}
    \caption{Accuracies with and without adversarial training on pairs of one source and one target speaker, on the development set.}
    \label{fig:OverviewPairs}
  \end{figure}%
}

\newcommand{\BaselineSingleDevTable}[1]{%
\begin{table}[#1]
\scriptsize
  \caption{Baseline word accuracies on \textit{single} speakers, averaged over the development set, with standard deviation. Layer types are \textbf{FC} (fully connected feedforward), \textbf{DP} (Dropout),
and \textbf{LSTM}, followed by the number of neurons/cells. $\ast$ marks the (reimplemented and recomputed) best system from \cite{Wand_ICASSP16_LipreadingLSTM}.}
  \label{tab:singlespk}
  \centering
  \begin{tabular}{p{3.4cm}cc}
    \toprule
	\textbf{Network} & \textbf{Training acc.} & \textbf{Test acc.} \\
    \midrule
	FC128-LSTM128-LSTM128 $\ast$ & 100.0\% \tpm 0.0\% & 78.5\% \tpm 5.6\% \\
	FC128-FC128-LSTM128 & 100.0\%  \tpm 0.0\%& 79.5\% \tpm 5.8\% \\
	FC256-FC256-LSTM256 & 100.0\%  \tpm 0.0\%& 79.4\% \tpm 5.7\% \\
	FC256-FC256-FC256-LSTM256 & 100.0\%  \tpm 0.0\%& 79.0\% \tpm 5.6\% \\
    \midrule
	FC256-DP-FC256-DP-FC256-DP-LSTM256 & 96.4\% \tpm 1.9\% & 83.3\% \tpm 5.7\% \\
    \bottomrule
  \end{tabular}
\end{table}%
}

\newcommand{\BaselineMultiDevTable}[1]{%
\begin{table}[#1]
\scriptsize
  \caption{Baseline word accuracies on training \textit{across} speakers without adaptation by domain-adversarial training, averaged over the development set, with standard deviation. 
The best network from table \ref{tab:singlespk} (FC256-DP-FC256-DP-FC256-DP-LSTM256) was used.}
  \label{tab:multispk}
  \centering
  \begin{tabular}{cccc}
    \toprule
	\textbf{Number of} & \multicolumn{2}{c}{\textbf{Source spk}} & \textbf{Target spk} \\
	\textbf{training spk} & \textbf{Train acc.} & \textbf{Test acc.} & \textbf{Test acc.} \\
    \midrule
	1 & 96.6\% \tpm 1.5\% & 81.9\% \tpm 6.4\%  & 13.5\% \tpm 6.9\% \\
	4 & 89.3\% \tpm 1.9\% & 78.4\% \tpm 3.2\%  & 31.2\% \tpm 8.0\% \\
	8 & 82.1\% \tpm 0.8\% & 74.5\% \tpm 1.0\%  & 37.8\% \tpm 9.8\% \\
    \bottomrule
  \end{tabular}
\end{table}%
}

\newcommand{\BasicAdversarialAlternative}[1]{%
\begin{table}[#1]
\scriptsize
  \caption{Word accuracies and standard deviations for systems with adversarial training on \emph{all} target sequences
or a \emph{subset of 50} target sequences, on the development speakers.}
  \label{tab:adversarial_alt}
  \centering
  \begin{tabular}{cccc}
    \toprule
	\textbf{Adversarial} & \textbf{Number of}    & \textbf{Target}  & \textbf{Relative} \\
	\textbf{Training on} & \textbf{training spk} & \textbf{Test acc.} & \textbf{Improvement} \\
    \midrule
	\multirowcell{3}{None} 
                           & 1  & 13.5\% \tpm 6.9\% & - \\
	                   & 4  & 31.2\% \tpm 8.0\% & - \\
	                   & 8  & 37.8\% \tpm 9.8\% & - \\
    \midrule
	\multirowcell{3}{All Target Sequences} 
                           & 1  & 19.2\% \tpm 10.0\% & 42.0\% \\
	                   & 4  & 35.3\% \tpm 10.5\% & 13.1\% \\
	                   & 8  & 42.4\% \tpm 13.0\% & 12.2\% \\
    \midrule
	\multirowcell{3}{50 Target Sequences} 
                           & 1  & 18.9\% \tpm 8.9\% & 40.0\% \\
	                   & 4  & 32.9\% \tpm 10.5\% & 5.4\% \\
	                   & 8  & 41.8\% \tpm 9.7\% & 10.6\% \\
    \bottomrule
  \end{tabular}
\end{table}%
}

\newcommand{\EvalAdversarial}[1]{%
\begin{table}[#1]
\scriptsize
  \caption{Word accuracies, relative improvements, and p-values for systems with adversarial training, on the evaluation speakers. Significant results are marked with $\ast$.}
  \label{tab:eval_adversarial_alt}
  \centering
  \begin{tabular}{m{1.5cm}cccc}
    \toprule
	\textbf{Adversarial} & \textbf{Number of}    & \textbf{Target}    & \textbf{Relative}    & \multirow{2}{*}{\textbf{p-value}}  \\
	\textbf{Training on} & \textbf{training spk} & \textbf{Test acc.} & \textbf{Improvement} &  \\
    \midrule
	\multirowcell{3}{None} 
                           & 1  & 18.7\% & - & - \\
	                   & 4  & 39.4\% & - & - \\
	                   & 8  & 46.5\% & - & - \\
    \midrule
	\multirowcell{3}{All Target\\Sequences} 
                           & 1  & 25.4\% & 35.8\% & 0.0030$\ast$ \\
	                   & 4  & 43.6\% & 10.7\% & 0.0261$\ast$ \\
	                   & 8  & 49.3\% & 6.0\% & 0.0266$\ast$ \\
    \midrule
	\multirowcell{3}{50 Target\\Sequences} 
                           & 1  & 24.1\% & 28.9\% & 0.0045$\ast$ \\
	                   & 4  & 41.5\% & 5.3\% & 0.1367 \\
	                   & 8  & 47.0\% & 1.1\% & 0.3555 \\
    \bottomrule
  \end{tabular}
\end{table}%
}

\begin{document}

\maketitle
\begin{abstract}
We present a \textit{Lipreading} system, \ie a speech recognition system using only visual features, which uses \textit{domain-adversarial training}
for speaker independence. Domain-adversarial training is integrated into the optimization  of a lipreader based on a stack of feedforward and
LSTM (Long Short-Term Memory) 
recurrent neural networks, yielding an end-to-end trainable system which only requires a very small number of frames of \textit{untranscribed} target data
to substantially improve the recognition accuracy on the target speaker. On pairs of different source and target speakers, we achieve a relative accuracy improvement of around 40\% 
with only 15 to 20 seconds of untranscribed target speech data. On multi-speaker training setups, the accuracy improvements are smaller but 
still substantial.
\end{abstract}

\section{Introduction}
\label{sec:intro}
\textit{Lipreading} is the process of understanding speech by using solely visual features, \ie images of the lips of a speaker.
In communication between humans, lipreading has a twofold relevance \cite{Woodhouse_ReviewVisualSpeechPerception}: 
First, visual cues play a role in spoken conversation \cite{McGurk_HearingLips}; second, hearing-impaired persons may use lipreading
as a means to follow verbal speech.

With the success of computer-based speech recognition over the past decades, automatic lipreading has become an active field of research as well,
with pioneering work by Petajan \cite{Petajan_PhD_Lipreading}, who used lipreading to augment conventional acoustic speech 
recognition, and Chiou and Hwang \cite{Chiou_LipreadingFromColorVideo}, who were the first to perform lipreading without resorting to any acoustic signal at all.
Since 2014, lipreading systems have systematically begun to use neural networks at part of the processing pipeline \cite{Noda_IS14_LipreadingUsingConvNet,Petridis_ICASSP16_DeepBottleneckFeaturesVisualASR} or 
for end-to-end-training  \cite{Wand_ICASSP16_LipreadingLSTM,Assael_LipNet,Chung_LipreadingInTheWild}. In our previous work \cite{Wand_ICASSP16_LipreadingLSTM},
we proposed a fully neural network based system, using a stack 
of fully connected and recurrent (LSTM, Long Short-Term Memory) \cite{Hochreiter_LongShortTermMemory,Gers_LearningToForget}
neural network layers. 

The scope of this paper is the introduction of state-of-the-art methods for \textit{speaker-independent lipreading} with neural networks. We evaluate our established system
\cite{Wand_ICASSP16_LipreadingLSTM} in a cross-speaker setting, observing a drastic performance drop on unknown speakers.
In order to alleviate the discrepancy between training speakers and  unknown test speaker, 
we use \textit{domain-adversarial training} as proposed by Ganin and Lempitsky \cite{Ganin_ICML15_UnsupervisedDomainAdaptationBackprop}:
\textit{Untranscribed} data from the target speaker is used as additional training input to the neural network, with the aim of pushing the network to learn an intermediate data representation
which is domain-agnostic, \ie which does not depend on whether the input data comes from a source speaker or a target speaker. We evaluate our system on a subset 
of the GRID corpus \cite{Cooke_GRIDCorpus}, which contains extensive data from 34 speakers and is therefore ideal for a systematic evaluation of
the proposed method.

\section{Related work}
\label{sec:related}

Lipreading can be used 
to complement or augment speech recognition, particularly in the presence of noise \cite{Petajan_PhD_Lipreading,Abdelaziz_DynamicWeightsCoupledAVSR},
and for purely visual speech recognition \cite{Chiou_LipreadingFromColorVideo,Bowden_SPIE13_RecentDevelopmentsLipreading,Noda_IS14_LipreadingUsingConvNet}.
In the latter case, ambiguities due to incomplete information (\eg about voicing) can be mitigated
by augmenting the video stream with \textit{ultrasound} images of the vocal tract \cite{Hueber_SSIUltrasound}.
Visual speech processing is an instance of a \textit{Silent Speech interface} \cite{Denby2010}; further promising approaches 
include  capturing the movement of the articulators by electric or permanent magnetic articulography \cite{Wang_IS15_SpeakerIndependentSSR,Gonzalez_SilentSpeechPMADirectSynthesis}, and 
capturing of muscle activity using electromyography \cite{Wand_EMGSpeakingModeVarieties,Wand_IS14_RealLifeUnsupAdapt,Deng_IS14_PracticalSSRSystem,Wand_IS16_DNNFrontendEMG}. 

Versatile lipreading features have been proposed, such as
Active Appearance Models \cite{Matthews_VisualFeaturesLipreading}, Local Binary Patterns \cite{Zhao_LipreadingSpatiotemporalDescriptors}, and 
PCA-based \textit{Eigenlips} \cite{Bregler_ICASSP94_Eigenlips} and \textit{Eigentongues} \cite{Hueber_ICASSP07_EigentongueUltrasound}.
For tackling speaker dependency, diverse scaling and normalization techniques have been employed \cite{Cox_AVSP08_ChallengeMultispeakerLipreading,Lan_AVSP10_ImprovingVisualFeaturesLipReading}.
Classification is often done with Hidden Markov Models (HMMs), \eg 
\cite{Hueber_IS07_UltrasoundOpticalContinuousPhoneReco,Bowden_SPIE13_RecentDevelopmentsLipreading,Busso_IS14_LipreadingIsolatedDigits,Lan_AVSP09_VisualFeaturesLipreading}.
Mouth tracking is done as a preprocessing step \cite{Lan_AVSP09_VisualFeaturesLipreading,Bowden_SPIE13_RecentDevelopmentsLipreading,Noda_IS14_LipreadingUsingConvNet}.
For a comprehensive review see \cite{Zhou_VisualSpeechDecodingReview}.

Neural networks have early been applied to the Lipreading task \cite{Wolff_NIPS93_LipreadingByNeuralNetworks}, however, they have become widespread
only in recent years, with the advent of state-of-the-art learning techniques (and the necessary hardware).
The first deep neural network for lipreading was a seven-layer convolutional net 
as a preprocessing stage for an HMM-based word recognizer \cite{Noda_IS14_LipreadingUsingConvNet}. Since then, several end-to-end trainable systems were presented 
\cite{Wand_ICASSP16_LipreadingLSTM,Assael_LipNet,Chung_LipreadingInTheWild}. The current state-of-the-art accuracy on the GRID corpus is 
3.3\% error \cite{Chung_LipreadingInTheWild} using a very large set of \textit{additional} training data; so their result is not directly comparable to ours.

In \textit{domain adaptation}, it is assumed that a learning task exhibits a 
domain shift between the training (or \textit{source}) and test (or \textit{target}) data. This can be mitigated in several ways \cite{Pan_SurveyTransferLearning};
we apply 
\textit{domain-adversarial training} \cite{Ganin_ICML15_UnsupervisedDomainAdaptationBackprop}, where an intermediate layer in a multi-layer network
is driven to learn a representation of the input data which is optimized to be domain-agnostic, \ie to make it difficult to detect whether an input sample
is from the source or the target domain. A great advantage of this approach is the end-to-end trainability of the entire system. For a summary of
further approaches to domain adaptation with neural networks, we refer to the excellent overview in \cite{Ganin_ICML15_UnsupervisedDomainAdaptationBackprop}.

%

\section{Data and preprocessing}
\label{sec:data}

\GRIDPictureExamples{t}

We follow the data preprocessing protocol from \cite{Wand_ICASSP16_LipreadingLSTM}. 
We use the GRID corpus \cite{Cooke_GRIDCorpus}, 
which consists of video and audio recordings of $34$ speakers (which we name \textit{s1} to \textit{s34}) saying $1000$ sentences each.
All sentences have a fixed structure: \textit{command(4) + color(4) + preposition(4) + letter(25) + digit(10) + adverb(4)}, for example ``Place red at J 2, please'', where
the number of alternative words is given in parentheses. 
There are $51$ distinct words;
alternatives are randomly distributed so that context cannot be used for classification.
Each sentence has a length of 3 seconds at 25 frames per second, so the total
data per speaker is $3000$ seconds ($50$ minutes).
Using the annotations contained in the corpus, we segmented all videos at word level, yielding $6000$ word samples per speaker.

We experiment on speakers s1--s19: speakers 1-9 form the \textit{development} speakers, used to determine optimal parameters;
speakers 10--19 are the \textit{evaluation} speakers, held back until the final evaluation of the systems.
The data from each speaker was randomly subdivided into training, validation, and test sets, 
where the latter two contain five samples of each word, \ie a total of $51 \cdot 5 = 255$ samples each.
The training data is consequently highly unbalanced: For example, each letter from ``a'' to ``z'' appears 30 times, whereas each color appears 240 times.

We converted the ``normal'' quality videos ($360\times 288$ pixels) to greyscale and extracted $40\times 40$ pixel windows containing the mouth area,
as described in \cite{Wand_ICASSP16_LipreadingLSTM}. The frames were contrast-normalized and z-normalized over the training set, 
\textit{independently} for each speaker. 
Unreadable videos were discarded. 

All experiments have one dedicated \textit{target} speaker on which this experiment is evaluated, and one, four, or eight \textit{source}
speakers on which supervised training is performed. Speakers are chosen consecutively,
for example, the experiments on \textit{four} training speakers on the development data are 
(s1 $\ldots$ s4) $\rightarrow$ s5,
(s2 $\ldots$ s5) $\rightarrow$ s6, $\cdots$,
(s9, s1, s2, s3) $\rightarrow$ s4, where $\rightarrow$ separates source and target speakers.
%
We also compute baseline results on single speakers. The data sets of each speaker are used as follows: \textit{Training} data is used for  
supervised training (on the source speakers) and unsupervised adaptation (on the target speaker). \textit{Validation} data is used for early stopping,
the network is evaluated on the \textit{test} data.

\section{Methods and System Setup}
\label{sec:methods}

\begin{figure}[t]
  \centering
  \includegraphics[width=.6\linewidth]{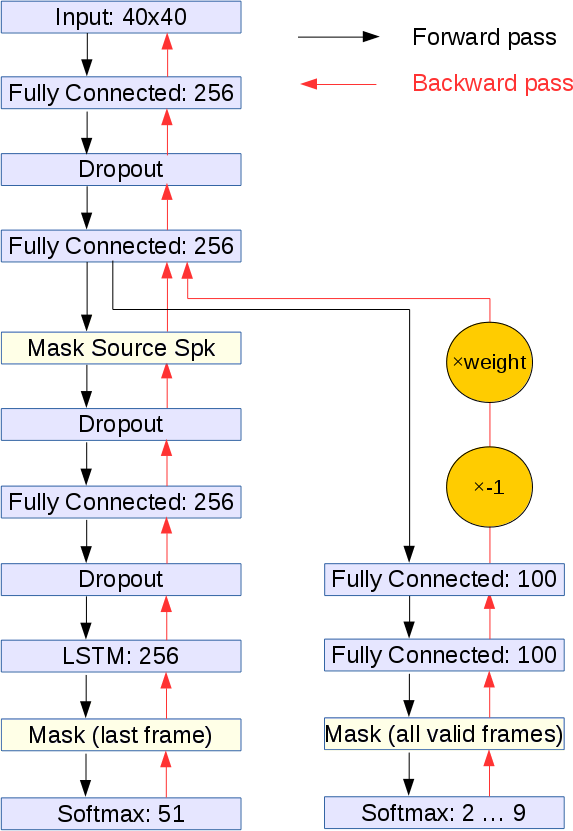}
  \caption{Optimal network topology for adversarial training, with a common part (top), word classifier (bottom left, only on source speakers), and speaker 
classifier (bottom right). Note that the gradient of the speaker classifier is inverted, and that the contribution of the adversarial network
is configurable.}
  \label{fig:adversarialsys}
\end{figure}

The system is based on the lipreading setup from \cite{Wand_ICASSP16_LipreadingLSTM}, reimplemented
in Tensorflow \cite{tensorflow_whitepaper_short}. Raw $40\times 40$ lip images are used as input data, \textit{without} any further preprocessing except normalization.
We stack several fully connected feedforward layers, optionally followed by Dropout \cite{Hinton_Dropout},  and one LSTM recurrent layer to form a network which is capable of recognizing sequential video data. 
The final layer is a softmax with 51 word targets. All inner layers use a tanh nonlinearity.
During testing, classification is performed on the \textit{last} frame of an input word, the softmax output on all previous frames is discarded. Similarly,
during training, an error signal is backpropagated (through time and through the stack of layers) only from the last frame of each training word sample.

Optimization is performed by minimizing the multi-class cross-entropy using stochastic gradient descent 
applying Tensorflow's \texttt{MomentumOptimizer} with a 
momentum of 0.5, a learning rate of 0.001, and a batch size of 8 sequences. The network weights are initialized following a truncated normal distribution with a standard deviation of 0.1.
In order to compensate for the unbalanced training set, 
each training sample is weighted with a factor inversely proportional to its frequency. 
Early stopping (with a patience of 30 epochs) is performed on the validation data
of the \textit{source} speakers.

\textbf{Adversarial training} \cite{Ganin_ICML15_UnsupervisedDomainAdaptationBackprop} is integrated 
as follows. At the second feedforward layer, we attach a further network which performs framewise \textit{speaker}
classification on source and target speakers. For this purpose, each training batch of 8 word sequences is augmented by eight additional
word sequences from the target speaker, for which \textit{no word label is used}, and \textit{no gradient is backpropagated} from the word classifier.
On the extended batch of 16 sequences, the ``adversarial'' network performs framewise speaker classification. This 
network follows a standard pattern (two feedforward layers with 100 neurons each plus a softmax layer with 2, 5, or 9 speaker outputs)
and is trained jointly with the word classifier, with a configurable weight. If there are more word sequences from the source
speaker(s) than from the target speaker, target sequences are repeated.

So far, this describes a joint classifier for two different tasks (speaker and word classification), resembling Caruana's Multitask training \cite{caruana1997}.
The power of the adversarial network comes from a simple twist: The backpropagated gradient from the adversarial network is \textit{inverted}
where it is fed into the main branch of the network, causing the lower branch to perform gradient \textit{ascent} instead of descent.
Since the speaker classification part of the system learns to classify speakers, the inverted gradient fed into the ``branching'' layer causes the
joint part of the network to learn to \textit{confuse} speakers instead of separating them. 
The speaker classifier and the joint network
 work for opposite objectives (hence, ``adversarial''); an idea first presented in the context of \textit{factorial codes} \cite{Schmidhuber_FactorialCodes}.
Figure \ref{fig:adversarialsys} shows a graphical overview of the system: The joint part is  at the top, at the bottom are
word classifier (left) and speaker classifier (right). 

\section{Experiments and Results}
\label{sec:experiments}

\subsection{Baseline Lipreader}
\label{ss:baseline}

\BaselineSingleDevTable{t}

The first experiment deals with establishing a baseline for our experiments, building on prior work  \cite{Wand_ICASSP16_LipreadingLSTM}. We 
run the lipreader as a \textit{single-speaker} system with different topologies,
optionally using Dropout (always with 50\% dropout ratio) to avoid overfitting the training set. Adversarial training is not used (\ie the weight in
figure \ref{fig:adversarialsys} is set to zero). 
Table \ref{tab:singlespk} shows the resulting test set accuracies averaged over the development speakers.

Without using Dropout, the accuracy on the test set is $\sim$79\%. Note in particular that the 
baseline
cannot substantially be improved by increasing the layer size or adding more layers. We remark that not only the average accuracy across speakers, but also
the accuracies for every single speaker hardly vary.

The situation changes when Dropout is used: Now our best average accuracy is 83.3\%, which is in line with 
results reported in literature (the most recent best result is 86.4\% word accuracy \cite{Gergen_IS16_TurboAudiovisualASR}, but with a different training/test data split). 
This best system, which is employed in the remainder of this paper, uses three feedforward layers each followed by Dropout, with 256 neurons each, followed by the LSTM layer with 256 LSTM cells, 
and the softmax layer. Thus the system is larger and has more layers than the baseline system, which is indeed made possible by the Dropout regularizer.

On the evaluation speakers, the baseline system achieves an average accuracy of 78.3\%, and the Dropout system
is at 83.9\% accuracy. This improvement is significant (one-tailed t-test with paired samples, $p=2.38 \times 10^{-6}$).


The accuracies in a cross-speaker setting, again on the development speakers, are given in table \ref{tab:multispk}. The accuracy decreases drastically, in particular when
only one source speaker is used for training: On an unknown target speaker, the system achieves only an average 13.5\% accuracy. The situation is clearly better
when training data from multiple speakers is used, but even for eight training speakers, the average accuracy on an unknown speaker is only 37.8\%. We also note
that the test accuracy on the \textit{source} speakers does not rise when data from multiple speakers is used, even though there is more training data.
It appears that the additional data does not ``help'' the system to improve its performance. On an unknown speaker, however, training data from multiple
speakers does improve performance, very probably the system learns to be more speaker-agnostic. A similar observation with a very different
input signal was reported in \cite{Wand_BS11_SessionIndEMG}.

\BaselineMultiDevTable{t}

Clearly, lipreading across different speakers is a challenging problem. In the remainder of this paper, we show how domain-adversarial training helps to tackle this
challenge.

\subsection{Tuning of the Adversarial System}

\BasicAdversarialAlternative{b}

We now augment the baseline word classification network with adversarial training as described in section \ref{sec:methods}, thus making full use of the 
system shown in figure \ref{fig:adversarialsys}.  For now, we use \textit{all} sequences from the training set of the target speaker.
As suggested in \cite{Ganin_ICML15_UnsupervisedDomainAdaptationBackprop}, we found it beneficial to gradually activate adversarial training: 
the weight of the adversarial part is set to zero at the beginning, every 10 epochs, it is raised by 0.2 until the maximum value of 1.0 is reached at epoch 50.
The results of this experiment are shown in the upper two blocks of table \ref{tab:adversarial_alt}, where it can be seen that adversarial training causes
substantial accuracy improvement, particularly with only one source speaker: In this case, the accuracy rises by more than 40\% relative,
from 13.5\% to 19.2\%. In the case of four or eight source speakers, the accuracy improves by 13.1\% resp.~12.2\% relative.
We tuned this system using various topologies for the adversarial part, as well as different weight schedules  for adversarial training, finding
rather consistent behavior. The only setting which is emphatically discouraged is starting with an adversarial weight  greater than zero. See  section \ref{ss:analysis} for further analysis.

\subsection{Training with Very Little Target Data}

While the presented system does not require supervised training data from the target speaker, we still use the entire training set
of the target speaker. In practical applications, even unsupervised training data may only be sparsely available, so this
setup is somewhat undesired. 

Since  the content of the target training sequences is irrelevant for the adversarial training, we may hypothesize that we could also do with a much smaller set of
target training data. So as a final experiment, we reduce the number of training sequences for the target speaker. The training protocol remains as before;
in particular, training is always performed on the full set of source sequences, target sequences are repeated as necessary.

\OverviewPairs{t}

\EvalAdversarial{b}

The original number of 5490 target training sequences can be reduced to 50 sequences without a substantial loss of accuracy---this amounts
to only 15-20 seconds of untranscribed target data.
Results are shown in the 
lower block of table \ref{tab:adversarial_alt}: For example, in the case of a single source speaker, the target accuracy drops to 18.9\% instead of 19.2\%. 
The improvement
is lower when more source speakers are used. We hypothesize that this stems from the growing ratio between the number of source sequences and the number of target sequences. 

Finally, figure \ref{fig:OverviewPairs} shows an accuracy breakdown for speaker pairs, \ie for single-speaker supervised training.
In eight out of nine cases, domain-adversarial training clearly outperforms the baseline system, often by a substantial margin. We also observe that the accuracy gain 
depends very much on the speaker pair.

\subsection{Evaluation}

We evaluate our result on the evaluation speakers, \ie speakers 10--19 from the GRID corpus. The hypothesis to be tested states that adversarial training
improves the accuracy of the cross-speaker lipreader trained on one, four, or eight source speakers, using either all target sequences or 50 target sequences.
We use the one-tailed t-test with paired samples for evaluation.

Table \ref{tab:eval_adversarial_alt} shows the resulting accuracies, relative improvements, and p-values. Improvements are significant in all cases in which
the entire target speaker data is used. For 50 target sequences, significance can be ascertained only in the case of a single source speaker, but
we always get some improvement.

We finally note that when applying such a system in practice, untranscribed data is accrued \textit{continuously}:
so the quality of the system on the target speaker could be improved continuously as well, without requiring any extra data collection.

\section{Analysis}
\label{ss:analysis}

\CurvesWithAdv{t}

In this section we attempt to shed light on the effect of domain-adversarial training.
Figure \ref{fig:CurvesWithAdv} shows the progress of training for speakers s5 $\rightarrow$  s6 versus the training epoch, 
with adversarial training activated. The \textit{source speaker} accuracies on validation and test set are $\sim$78\%,
almost unaffected by adversarial training. The \textit{target speaker} accuracies are
39.1\% on the validation set and 39.5\% on the test set, our greatest single increase with adversarial training:
without adversarial training, the target accuracy is less than 22\%.

From the steady rise of the first curve, we see that the training progresses smoothly.
This is the expected behavior for a well-tuned system. 
On the \textit{validation} sets, the accuracy varies much less smoothly, with jumps of several percent points between epochs. 
We observed that this behavior is quite consistent for all systems, with or without adversarial training, and also for varying numbers
of training speakers. Clearly the ``error landscape'' between training and validation data is very different, both within the same speaker
and between different speakers.

The effect of adversarial training is clearly observable: At epoch 10, where
adversarial training becomes active (with 0.2 weight), the target accuracy jumps visibly, even though \textit{the criterion 
for which the adversarial network is optimized is very different from the word accuracy which is plotted in the graph}.
This is a remarkable success, even though it should be noted (compare figure \ref{fig:OverviewPairs}) that on other speaker 
pairs, we obtain a much lower improvement by adversarial training. 

\section{Conclusion}
\label{sec:conclusion}

In this study we have described how to apply \textit{domain-adversarial training}  to a state-of-the-art lipreading system for improved speaker independency.
When training and test are performed on pairs of different speakers, the average improvement is around 40\%, which is highly significant;
this improvement even persists when the amount  of untranscribed target data is drastically reduced to about 15-20 seconds. When supervised training
data from several speakers is available, there is still some improvement, from a much higher baseline

\section{Acknowledgements}
The first author was supported by the H2020 project INPUT (grant \#687795).

\bibliographystyle{IEEEtran}

\bibliography{IsBib}

\end{document}